\documentclass[runningheads]{llncs}
\usepackage[T1]{fontenc}
\usepackage{graphicx}

\usepackage{amssymb}
\usepackage{amstext}
\usepackage{amsmath}
\usepackage{amsfonts}       

\usepackage{url}

\usepackage{listings}
\usepackage{algorithm}
\usepackage{algorithmic}
\begin{document}
\title{Elements of Active Continuous Learning and Uncertainty Self-Awareness: a Narrow Implementation for Face and Facial Expression Recognition}
\titlerunning{Elements of Active Continuous Learning and Uncertainty Self-Awareness}
%
\author{Stanislav Selitskiy \orcidID{0000-0003-1758-0171} }
\authorrunning{S. Selitskiy}
%
\institute{School of Computer Science and Technology, University of Bedfordshire, Park Square, Luton, LU1 3JU, UK,\\ 
\email{stanislav.selitskiy@study.beds.ac.uk}\\
}
\maketitle              
\begin{abstract}
Reflection on one's thought process and making corrections to it if there exists dissatisfaction in its performance is, perhaps, one of the essential traits of intelligence. However, such high-level abstract concepts mandatory for Artificial General Intelligence can be modelled even at the low level of narrow Machine Learning algorithms. Here, we present the self-awareness mechanism emulation in the form of a supervising artificial neural network (ANN) observing patterns in activations of another underlying ANN in a search for indications of the high uncertainty of the underlying ANN and, therefore, the trustworthiness of its predictions. The underlying ANN is a convolutional neural network (CNN) ensemble employed for face recognition and facial expression tasks. The self-awareness ANN has a memory region where its past performance information is stored, and its learnable parameters are adjusted during the training to optimize the performance. The trustworthiness verdict triggers the active learning mode, giving elements of agency to the machine learning algorithm that asks for human help in high uncertainty and confusion conditions.

\end{abstract}
\section{Introduction}
\label{sec:in}

Artificial Intelligence (AI) is quite a vague terminology artefact that has been overused many times, sometimes even for describing narrow software implementations of simple mathematical concepts such as multi-dimensional regression. It is understandable that to separate the high-level AI from the narrow level, such abbreviation as AGI (Artificial General Intelligence) was introduced. Sometimes even AGI gets associated with the ``hype-style'' conversation, therefore such alternatives as ``human-level AI'' \cite{lecun2022path} or ``DL System 2'' \cite{BengioNeurips2019Sys}, and others can used. 
The founders of AI research, such as A. Turing and J. McCarthy, who coined the very term AI, were sceptical about the worthiness of the attempts to answer what AI is. Instead, they suggested answering the question of how well AI can emulate or implement human-type intelligence \cite{TURING1950,mccarthy1981some}. N. Chomsky, in numerous lectures and publications (f.e. \cite{Chomsky1996PowersAP}), even more categorically elaborated that AI is a human linguistic concept rather than an independent phenomenon. 

Suppose we accept discussing AI in the context of human-likeliness. There still should be room for learning from simple and narrow machine learning (ML) algorithms if they could be used as ``building blocks'' and working approximations of human-like intelligence. In this work, we want to concentrate on two aspects of human-likeliness intelligence functionality: continuous lifetime learning and awareness of uncertainty.

Lifetime Learning (LTL) was introduced in the mid-'90s in the context of the robot learning process \cite{thrun1995lifelong}. 
The LTL learner could face various tasks during its lifetime, and each new learning task may benefit from the saved successful models and examples of data they were trained and applied to \cite{thrun1995learning}. 

From the human perspective, a high volume of training data forced on the learner is not a benefit. Rather abundance of potential data, from which the learner chooses a few characteristic examples and asks for the teacher's expert advice, is more desirable. A similar Active Learning (AL) approach expects an ML algorithm to ask an ``Oracle'' advice on selected un-labelled data \cite{marcheggiani2014experimental}, in particular when high uncertainty about a particular piece of data is occurred \cite{lewis1994heterogeneous}. 

The idea of learning the ML processes was also introduced in the '90s by the same author as LTL \cite{Thrun98}, and recently gained traction in various flavours of meta-learning \cite{vanschoren2018metalearning}. One of the directions of learning about learning is learning uncertainty of the learner \cite{li2021mural}.


To bring general considerations into a practical, although narrow perspective, we concentrate on making the meta-learning supervisor ANN model. It learns patterns of the functionality of the underlying CNN models that are associated with the failed predictions for Face Recognition (FR) \cite{SelitskiyLOD2022} and Facial Expression Recognition (FER) tasks \cite{SelitskiyAIVR2021}, self-adjusting on the previous experience during training, as well as, test times.

The reason to use FR and FER tasks is based not only on the fact that these are pretty human-centric ones but also, although State of the art (SOTA) CNN models had already passed the milestone of the human-level accuracy of face recognition several years ago in the ideal laboratory conditions. In the case of the Out of (training) Data Distribution (OOD) \cite{qiu2021resisting}, for example, makeup and occlusions, accuracy significantly drops. Even worse for FER algorithms and modes, which perform far worse than FR. The reason may be that the idea that the whole spectre of emotion expressions can be reduced to six basic facial feature complexes \cite{ekman1971constants} was challenged because human emotion recognition is context-based. The same facial feature complexes may be interpreted differently depending on the situational context \cite{cacioppo2000psychophysiology}.

Applying the continuous uncertainty and trustworthiness self-awareness algorithms to FR and FER models and data sets built and partitioned to exaggerate and aggravate OOD conditions is a good area for evaluating the algorithms.


The paper is organized as follows. Section~\ref{sec:ps} proposes a solution for dynamically adjusting the meta-learning trustworthiness estimating algorithm for predictions done for the FR and FER tasks. Section \ref{sec:ds} describes the data set used for experiments; Section \ref{sec:ex} outlines experimental algorithms in detail; Section \ref{sec:res} presents the obtained results, and Section \ref{sec:dc} discusses the results, draws practical conclusions, and states directions of the research of not yet answered questions.

\section{Proposed Solution}
\label{sec:ps}

\subsection{Uncertainty Meta-learning}
In \cite{SelitskiyLOD2022}, two approaches to assigning a trustworthiness flag to the FR prediction were proposed: statistical analysis of the distributions of the maximal softmax activation value for correct and wrong verdicts, and use of the meta-learning supervisor ANN that uses the whole set of softmax activations for all FR classes (sorted into the ``uncertainty shape descriptor'' (USD) to provide class-invariant generalization) as an input and generates trusted or not-trusted flag. 

This contribution ``marries'' these two approaches by collecting statistical information about training results in the loss layer (LL) memory of the meta-learning supervisor ANN. The information in the LL's memory holds prediction result $y_t$, training label result $l_t$, and trustworthiness threshold $TT$. The latter parameter is the learnable one, and the derivative of the loss error, calculated from these statistical data, is used to auto-configure the TT to optimize the sum of square errors loss: $SSE_{TT} = \sum_{t=1}^{K} SE_t$, where $K$ is a number of entries in the memory table:

\begin{equation}
\label{eq:1}
SE_{TTt} = \begin{cases}
{(y_t - TT)}^2, & (l_t < TT \wedge \, y_t > TT) \vee \, (l_t > TT \wedge \, y_t < TT) \\
0, & (l_t > TT \wedge \, y_t > TT) \vee \, (l_t < TT \wedge \, y_t < TT)
\end{cases}
\end{equation}

\begin{figure}
\begin{minipage}[b]{1.0\textwidth}
  \centering
  \centerline{
  \includegraphics[width=0.33\linewidth]{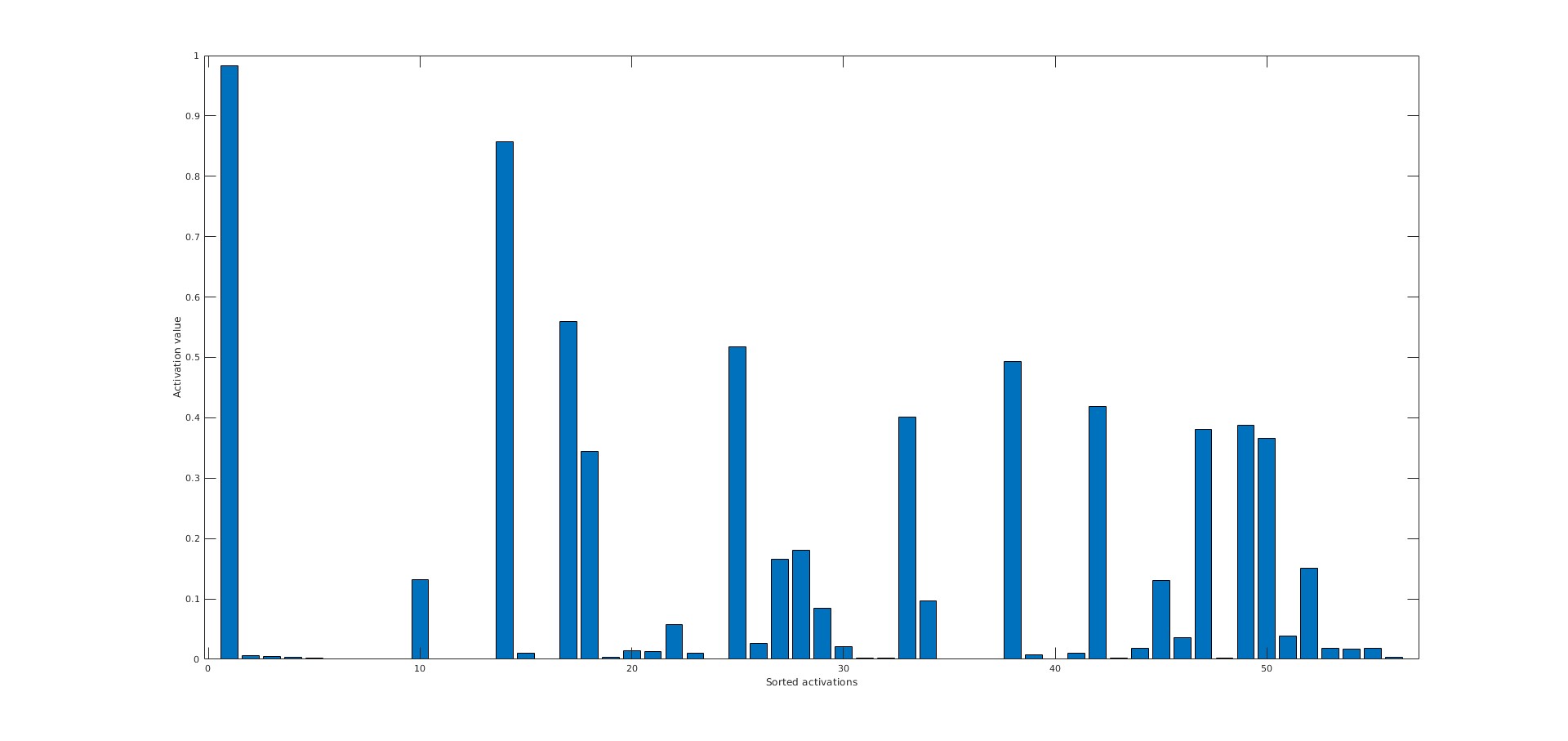}
  \includegraphics[width=0.33\linewidth]{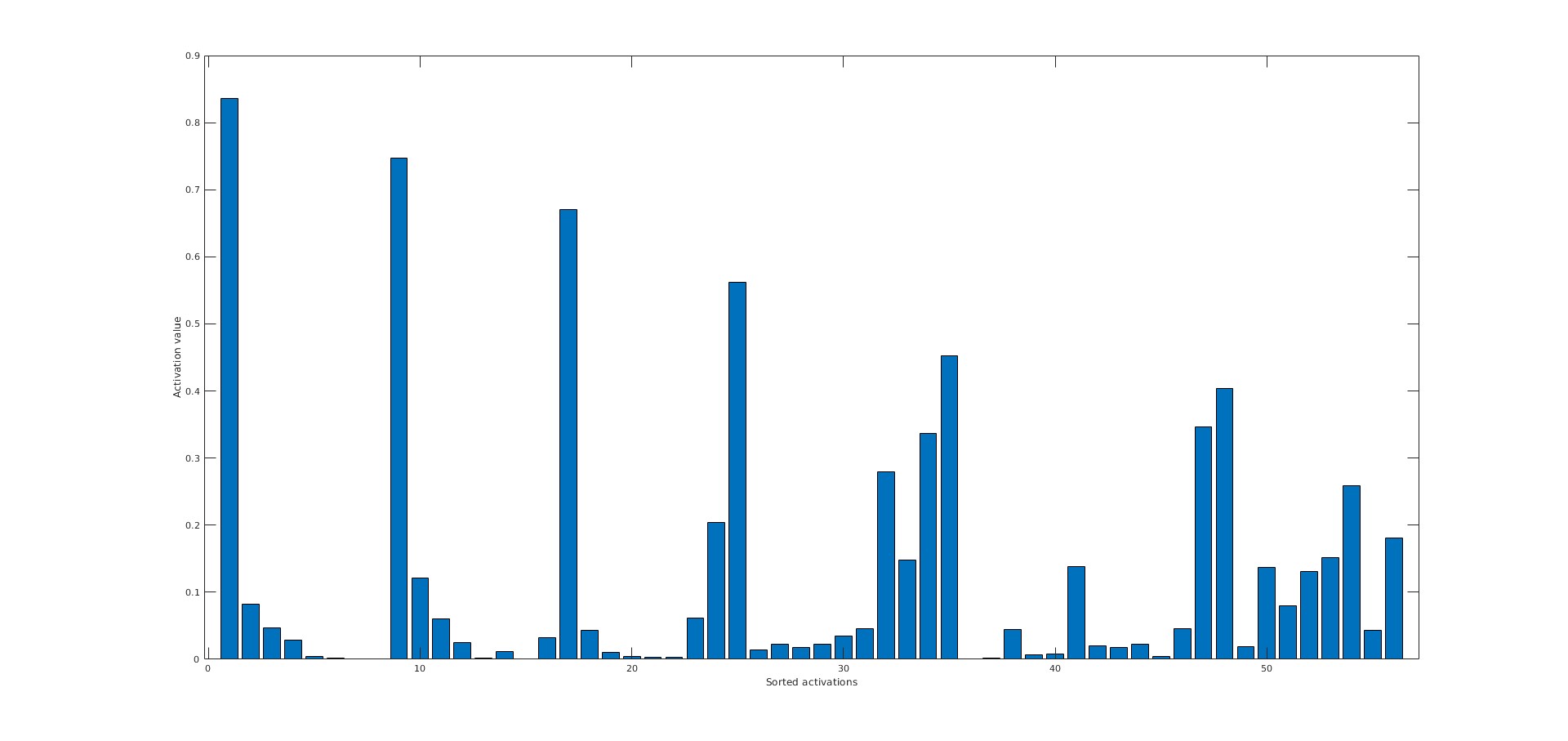}
  \includegraphics[width=0.33\linewidth]{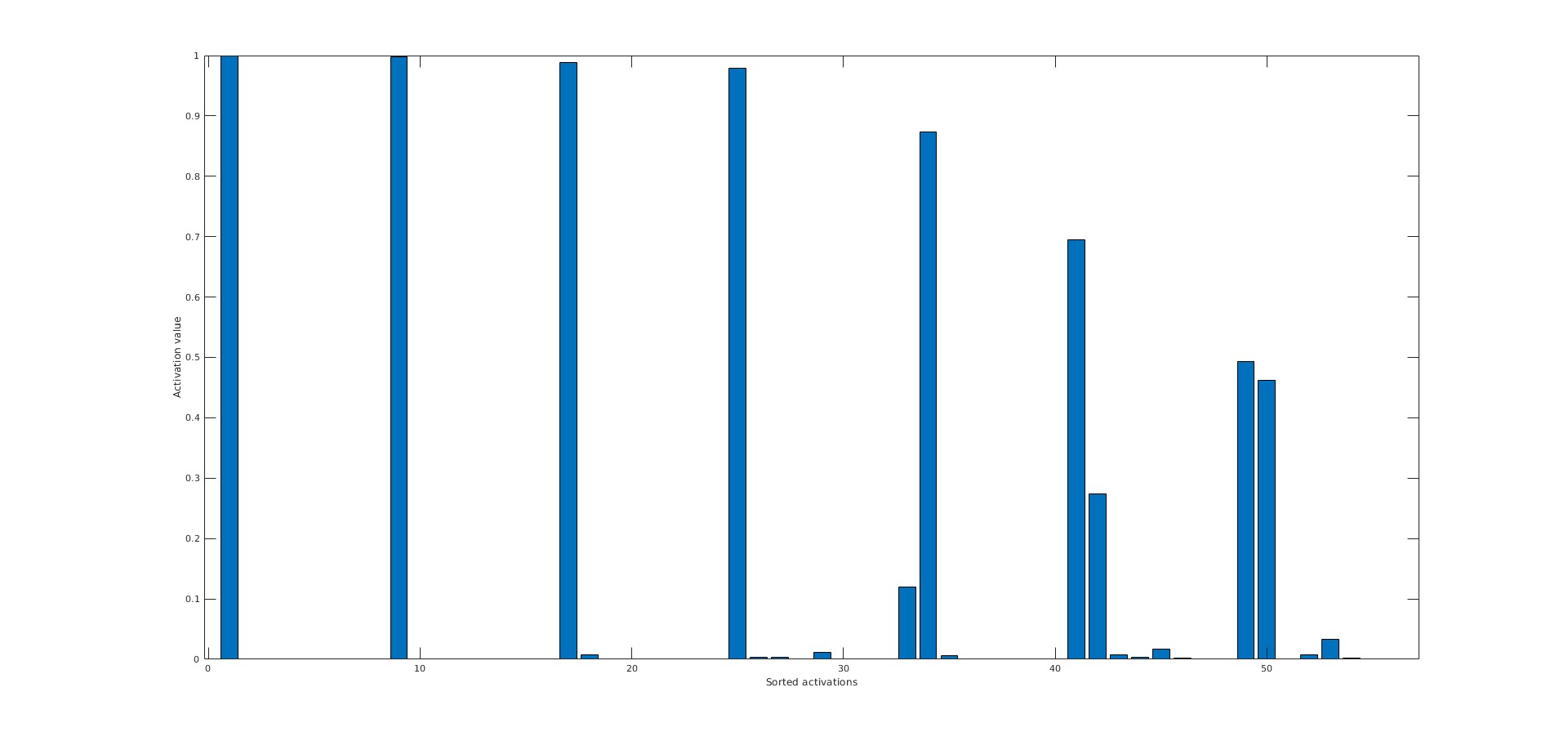}
  }
\end{minipage}
\caption{Examples of the uncertainty shape descriptors for 0, 4, and 6 correct CNN ensemble FER predictions.}
\label{fig:usd}
\end{figure}

The input of the meta-learning supervisor ANN was built from the softmax activations of the ensemble of the underlying CNN models. The algorithm of building USD can be described in a few words as follows: ``build the ``uncertainty shape descriptor'' by sorting softmax activations inside each model vector, order model vectors by the highest softmax activation, flatten the list of vectors, rearrange the order of activations in each vector to the order of activations in the vector with the highest softmax activation''.
Examples of the descriptor for the $M = 7$ CNN models in the underlying FR or FER ensemble, for the cases when none of the models, $4$, and $6$ detected the face correctly, are presented in Figure~\ref{fig:usd}. It could be seen that shapes of the distribution of the softmax activations are pretty distinct and, therefore, can be subject to the pattern recognition task performed by the meta-learning supervisor ANN.

\begin{figure}
\centering
  \centerline{
  \includegraphics[width=1.0\textwidth]{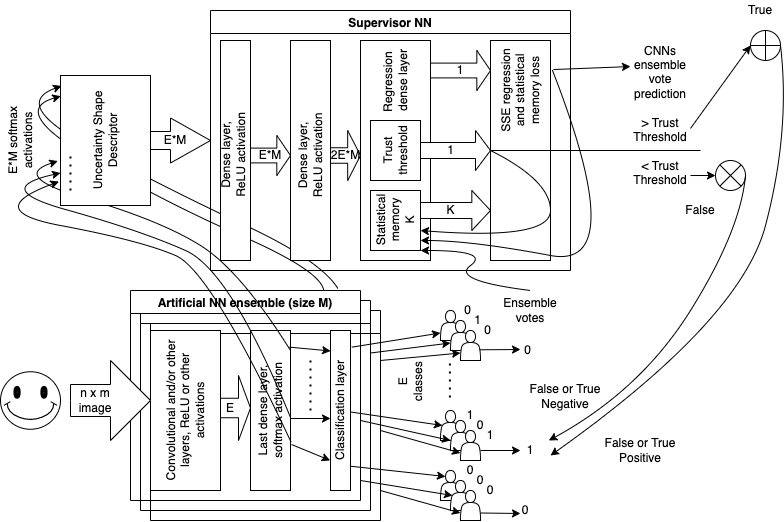}
  }
\caption{Meta-learning Supervisor ANN over underlying CNN ensemble.}
\label{fig:sann}
\end{figure}

However, unlike in the mentioned above publication, for simplification reasons, supervisor ANN was not categorizing the predicted number of the correct members of the underlying ensemble but instead was performing the regression task of the transformation. On the high level (ANN layer details are given in Section~\ref{sec:ex}), the transformation can be seen as Equation~\ref{eq:2}, where ${n = |C| * M}$ is the dimensionality of the $\forall \, USD \in \mathcal{X}$, $|C|$ - cardinality of the set of FR or FER categories (subjects or emotions), and $M$ - size of the CNN ensemble \ref{fig:sann}.

\begin{equation}
\label{eq:2}
reg:\mathcal{X}\subset \mathbb{R}^n \mapsto \mathcal{Y}\subset \mathbb{R}
\end{equation}
where $\forall \textbf{x} \in \mathcal{X} \, , \textbf{x} \in (0 \dots 1)^n \, , \forall y \in \mathcal{Y} \, , E(y) \in [0 \dots M]$.

The loss function used for $y$ is the usual for regression tasks, sum of squared error: $SSE_{y} = \sum_{t=1}^{N_{mb}} (y_j-e_j)^2$, where $e$ is the label (actual number of the members of CNN ensemble with correctl prediction), and $N_{mb}$ - minbatch size. 

From the trustworthiness categorization and ensemble vote point of view, the high-level transformation of the combined CNN ensemble together with the meta-learning supervisor ANN can be represented as Equation~\ref{eq:3}:

\begin{equation}
\label{eq:3}
cat:\mathcal{I}\subset \mathbb{I}^l \mapsto \mathcal{C}\subset \mathbb{C} \times \mathbb{B}
\end{equation}

where $\textbf{i}$ are images, $l$ - mage size, $c$ - classifications, and $b$ - binary trustworthy flags, such as $\forall \textbf{i} \in \mathcal{I} \, , \textbf{i} \in (0 \dots 255)^l \, , \forall c \in \mathcal{C} \, , c \in \{c_1, \dots, c_{|C|}\} \, , \forall b \in \mathcal{B} \, , b \in \{1, 0\}$.

\begin{equation}
\label{eq:4}
b_i = \begin{cases}
1, & \text{($y_i > TT_t$)} \\
0, & \text{($y_i < TT_t$)}
\end{cases}
\end{equation}

Where $i$ is an index of the image at the moment $t$ of the state of the loss function memory.

\begin{equation}
\label{eq:5}
c_i = argmin(|y_i-e_i(c_i)|)
\end{equation}

Equations above describe the ensemble vote that chooses category $c_i$, which received the closest number of votes $e_i$ to the predicted regression number $y_i$.

\subsection{Active CNN ensemble Learning and Life-time SNN Learning}
A common strategy for the passive ML algorithms is to detect OOD conditions \cite{williams2021fool}. AL paradigm expects actions upon such a condition discovery. In \cite{selitskaya2020challenges} we investigated the effects of the enrichment of the training data by a few examples of the OOD makeup and occlusion examples. Those enrichments were either ``expert'' driven or random. Here, the SNN verdict to detect confusion of the underlying CNN ensemble (i.e. when the SNN-predicted number of correct CNN models in the ensemble is less than the trusted threshold: $y_t < TT_t$) is used to invoke AL mode post-classification, asking ``Oracle'' to assign the correct label to the problematic image.

The trained CNN ensemble has a reference training set $|D_r| = N_{mb}$ of the mini-batch size, which is composed of the randomly-selected elements of the whole training set $D_{tr} ; D_r \subset D_{tr}$. Upon ``Oracle'' labelling, one of the elements of the older reference training set is replaced, and original CNN ensemble models are quickly retrained on a few epochs. The percentage of the allowed ``Oracle'' requests is limited by a low number.

Lifetime or (continuous or online) learning for SNN is implemented similarly to AL for CNN ensemble; however, because SNN has no supervisor of supervisor, retraining on the reference training set is implemented after each test classification.

\section{Data Set}
\label{sec:ds}
The BookClub artistic makeup data set contains images of $E=|C|=21$ subjects. Each subject's data may contain a photo-session series of photos with no makeup, various makeup, and images with other obstacles for facial recognition, such as wigs, glasses, jewellery, face masks, or various headdresses. The data set features $37$ photo sessions without makeup or occlusions, $40$ makeup sessions, and $17$ sessions with occlusions. Each photo session contains circa 168 JPEG images of the $1072 \times 712$ resolution of six basic emotional expressions (sadness, happiness, surprise, fear, anger, disgust), a neutral expression, and the closed eyes photoshoots taken with seven head rotations at three exposure times on the off-white background. The subjects' age varies from their twenties to sixties. The race of the subjects is predominately Caucasian and some Asian. Gender is approximately evenly split between sessions. 

The photos were taken over two months, and several subjects were posed at multiple sessions over several weeks in various clothing with changed hairstyles, downloadable from \url{https://data.mendeley.com/datasets/yfx9h649wz/3}. All subjects gave written consent to use their anonymous images in public scientific research. 


%
%
%

\section{Experiments}
\label{sec:ex}
The experiments were run on the Linux (Ubuntu 20.04.3 LTS) operating system with two dual Tesla K80 GPUs (with $2\times 12$GB GDDR5 memory each) and one QuadroPro K6000 (with $12$GB GDDR5 memory, as well), X299 chipset motherboard, 256 GB DDR4 RAM, and i9-10900X CPU. Experiments were run using MATLAB 2022a.

The experiments were done using MATLAB with Deep Learning Toolbox. For FR and FER experiments, the Inception v.3 CNN model was used. Out of the other SOTA models applied to FR and FER tasks on the BookClub data set (AlexNet, GoogLeNet, ResNet50, InceptionResnet v.2), Inception v.3 demonstrated overall the best result over such accuracy metrics as trusted accuracy, precision, and recall \cite{SelitskiyLOD2022,SelitskiyAIVR2021}. Therefore, the Inception v.3 model, which contains $315$ elementary layers, was used as an underlying CNN. Its last two layers were resized to match the number of classes in the BookClub data set ($21$), and retrained using ``adam'' learning algorithm with $0.001$ initial learning coefficient, ``piecewise'' learning rate drop schedule with $5$ iterations drop interval, and $0.9$ drop coefficient, mini-batch size $128$, and $10$ epochs parameters to ensure at least $95\%$ learning accuracy. The Inception v.3 CNN models were used as part of the ensemble with a number of models $N=7$ trained in parallel.

\begin{figure}
\begin{minipage}[b]{1.0\textwidth}
  \centering
  \centerline{
  \includegraphics[width=0.5\linewidth]{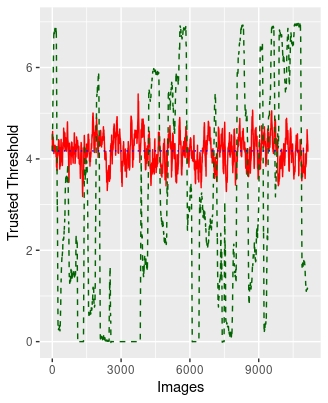}
  \includegraphics[width=0.5\linewidth]{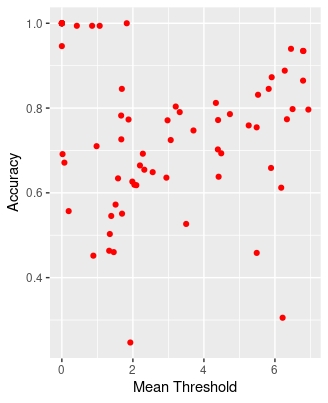}
  }
\end{minipage}
\caption{left - trusted threshold learned during the training phase (blue, dashed line), online learning changes for grouped test images (green), and shuffled test images (red) for FR task. Right - trusted accuracy against the trusted threshold for grouped test images for the FR task.}
\label{fig:cl}
\end{figure}

Meta-learning supervisor ANN models were trained using the ``adam'' learning algorithm with $0.01$ initial learning coefficient, mini-batch size $64$, and $200$ epochs. The memory buffer length, which collects statistics about previous training iterations, was set to $K=8192$.
For online learning SNN experiments, the number of epochs was set to $10$, and for active learning, the number of epochs was set to $5$. Limits for the ``Oracle'' requests were set for $1\%$ and $0.1\%$, resulting in $112$ and $12$ requests out of $11125$ test images.

The $reg$ meta-learning supervisor ANN transformation represented in the Equation~\ref{eq:2} implemented with two hidden layers with $n+1$ and $2n+1$ neurons in the first and second hidden layer, and ReLU activation function. All source code and detailed results are publicly available on GitHub \url{https://github.com/Selitskiy/StatLoss}.


\subsection{Trusted Accuracy Metrics}
\label{sec:tam}

Suppose only the classification verdict is used as a final result of the ANN model. In that case, the accuracy of the target CNN model can be calculated only as the ratio of the number of correctly identified test images by the CNN model to the number of all test images:

\begin{equation}
\label{eq:12}
Accuracy = \frac{N_{correct}}{N_{all}}
\end{equation}

When additional dimension in classification is used, for example amending verdict of the meta-learning supervisor ANN, (see Formula~\ref{eq:3}), and $cat (\textbf{i}) = c \times b$, where $\forall \textbf{i} \in \mathcal{I}, \, \forall c \times b \in \mathcal{C} \times \mathcal{B} = \{(c_1,b_1), \dots (c_p, b_p)\}, \, \forall b \in \mathbb{B} = \{True, False\}$, then the trusted accuracy and other trusted quality metrics can be calculated as:
\begin{equation}
\label{eq:13}
Accuracy_t = \frac{N_{correct:f = T} + N_{wrong:f \ne T}}{N_{all}}
\end{equation}
As a mapping to a more usual notations, $N_{correct:f = T}$ can be as the True Positive (TP) number, $N_{wrong:f \ne T}$ - True Negative (TN), $N_{wrong:f = T}$ - False Positive (FP), and $N_{correct:f \ne T}$ - False Negative (FN). 
Analogously to the trusted accuracy, we used metrics such as trusted precision, recall, specificity and F1 score for the models' evaluation.







%
%
%

\section{Results}
\label{sec:re}
Results of the FR experiments are presented in the Table~\ref{tab1}, for FER experiments - in the Table~\ref{tab2}. 
The first column holds accuracy metrics using the ensemble's maximum vote. The second column - using the ensemble vote closest to the meta-learning SNN prediction and trustworthiness threshold learned only on the test set, see Formulae~\ref{eq:4},\ref{eq:5}. The third column lists accuracy metrics in the continuous learning setting, and the next two - are in the active learning setting with $1\%$ and $0.1\%$ of the allowed ``Oracle'' requests.


Figure~\ref{fig:cl} on the left shows trustworthy thresholds learned during training and continuous learning setting when test data is either unstructured or structured by a photo session,  i.e. groups of the same person and same makeup or occlusion, but with different lighting, head position, and emotion expression.
Figure~\ref{fig:cl} on the right shows the relationship between the average session trusted threshold and session-specific trusted recognition accuracy for FR and FER cases of the grouped test sessions.


\begin{table}
\caption{Accuracy metrics for FR task. Maximal ensemble vote, SNN predicted vote, SNN with online retraining, CNN ensemble active learning on 1\% of test data, CNN ensemble active learning on 0.1\% of test data.}
\label{tab1}
\begin{tabular}{|l|l|l|l|l|l|}
\hline
Metric & Maximal & Predicted & Online & Active 1\% & Active 0.1\%\\
\hline
Untrusted accuracy & 0.68237 & 0.57488 & 0.57488 & 0.84720 & 0.60335\\
Trusted accuracy & 0.73836 & 0.83383 & 0.83526 & 0.92293 & 0.81833\\
Trusted precision & 0.84102 & 0.91644 & 0.91821 & 0.99198 & 0.9560\\
Trusted recall & 0.76029 & 0.78227 & 0.78321 & 0.91644 & 0.73257\\
Trusted F1 score & 0.79862 & 0.84406 & 0.84535 & 0.95271 & 0.82952\\
Trusted specificity & 0.69124 & 0.90355 & 0.90566 & 0.95890 & 0.94877\\
\hline
\end{tabular}
\end{table}

\begin{table}
\caption{Accuracy metrics for FER task. Maximal ensemble vote, SNN predicted vote, SNN with online retraining, CNN ensemble active learning on 1\% of test data.}
\label{tab2}
\begin{tabular}{|l|l|l|l|l|}
\hline
Metric & Maximal & Predicted & Online & Active 1\% \\
\hline
Untrusted accuracy & 0.39626 & 0.29147 & 0.29147 & 0.20126\\
Trusted accuracy & 0.672599 & 0.77678 & 0.69298 & 0.77735\\
Trusted precision & 0.64736 & 0.65266 & 0.48052 & 0.42908\\
Trusted recall & 0.38166 & 0.50052 & 0.65778 & 0.32157\\
Trusted F1 score & 0.48021 & 0.56656 & 0.55534 & 0.36763\\
Trusted specificity & 0.86355 & 0.89042 & 0.70747 & 0.89219\\
\hline
\end{tabular}
\end{table}

\section{Discussion, Conclusions, and Future Work}
\label{sec:dc}
For the experimentation with CNN model ensemble based on Inception v.3 architecture and data set with significant OOD in the form of makeup and occlusions, using meta-learning SNN, which works as an instrument of self-awareness of the model about uncertainty and trustworthiness of its predictions, noticeably increases accuracy metrics for FR tasks (by tens of per cent) and significantly (doubles) - for FER task. The proposed novel loss layer with memory architecture without online retraining increases key accuracy metrics by an additional (up to 5) percentage. The trustworthiness threshold learned using the loss layer with memory offers a simple explanation of why prediction for a given image was categorized as trusted or non-trusted.

Active learning significantly improves FR accuracy metrics even at the $0.1\%$ of the allowed ``Oracle'' requests and brings trusted accuracy metrics at the high $90\%$ level for $1\%$ of the allowed test-time requests. For FER task, AL significantly improves accuracy metrics related to true negatives.

\begin{figure}
\begin{minipage}[b]{1.0\textwidth}
  \centering
  \centerline{
    \includegraphics[width=0.5\linewidth]{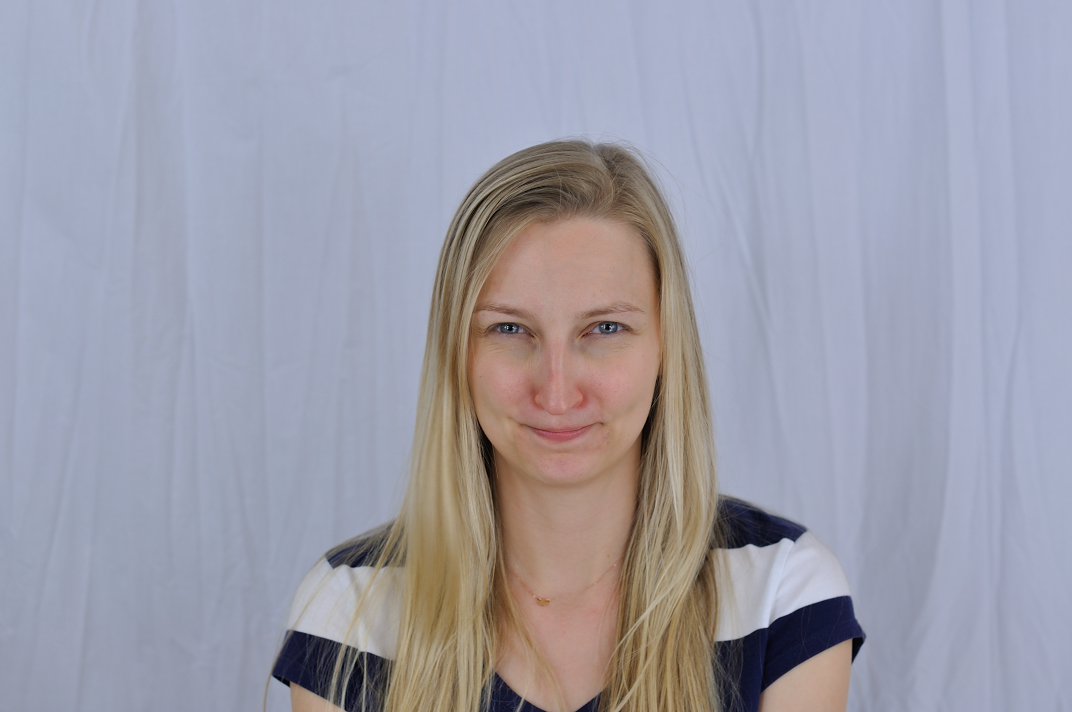}
    \includegraphics[width=0.5\linewidth]{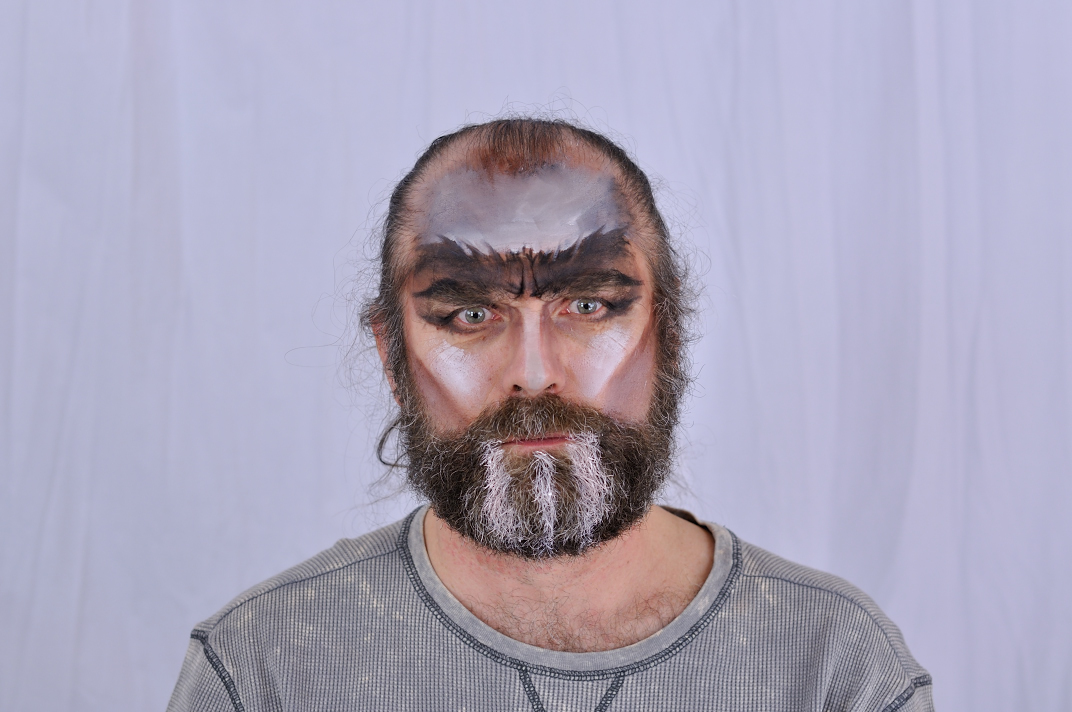}
  }
\end{minipage}
\caption{Examples of images for FER (anger expression) with the low trusted threshold (bad acting) - left and high trusted threshold (better acting) - right.}
\label{fig:im_th}
\end{figure}

Online retraining adds insignificant improvement to accuracy metrics for non-structured test data. However, online retraining of the trustworthiness threshold on structured test data informs the model not only about its uncertainty but also about the quality of the test session, see Figure~\ref{fig:im_th}. For example, it could be seen that a low-threshold session has a poorly performing subject who struggles to play the anger emotion expression, while in the high-threshold session, the facial expression is much more apparent.



\label{sec:res}
%
%
%


%
%
%
\bibliographystyle{splncs04}
\bibliography{refs,refs_vs}

\end{document}